\documentclass[11pt,a4paper]{article}
\usepackage[hyperref]{emnlp2018}
\usepackage{times}
\usepackage{latexsym}
\usepackage[utf8]{inputenc}
\usepackage{url}
\usepackage{graphicx}
\graphicspath{ {images/} }
\usepackage{amsfonts}
\usepackage{amsmath}
\usepackage{multirow}
\usepackage{makecell}
\usepackage{censor}
\usepackage{microtype}
\usepackage{amsmath,amsfonts,amssymb}
\usepackage{tabularx}
\usepackage{float}
\usepackage{wrapfig}

\usepackage{xcolor}
\definecolor{fl}{HTML}{00C8FF}
\definecolor{rz}{HTML}{800080}
\definecolor{Mycolor2}{HTML}{00F9DE}
\definecolor{ea}{HTML}{339933}
\definecolor{sc}{HTML}{800000}

\newcommand\kedznote[1]{}
\newcommand\fnote[1]{}
\newcommand\rznote[1]{}
\newcommand\kmnote[1]{}
\newcommand\svnote[1]{}
\newcommand\eanote[1]{}
\definecolor{sc}{HTML}{800000}
\newcommand\scnote[1]{}

\aclfinalcopy 

\title{Detecting Gang-Involved Escalation on Social Media Using Context}

\author{Serina Chang\textsuperscript{1}, 
  Ruiqi Zhong\textsuperscript{1}, 
  Ethan Adams\textsuperscript{1},
  Fei-Tzin Lee\textsuperscript{1},
  Siddharth Varia\textsuperscript{1}, \\
  \textbf{Desmond Patton\textsuperscript{2},
  William Frey\textsuperscript{2},
  Chris Kedzie\textsuperscript{1},
  and Kathleen McKeown\textsuperscript{1}} \\
  \textsuperscript{1}Department of Computer Science, Columbia University \\
  \textsuperscript{2}School of Social Work, Columbia University\\\AND
  {\tt \{sc3003, rz2383, ea2678, fl2301, sv2504, dp2787, w.frey\}@columbia.edu} \\
  {\tt \{kedzie, kathy\}@cs.columbia.edu}
  }

\date{}
 
\begin{document}
\maketitle


\begin{abstract}
Gang-involved youth in cities such as Chicago have increasingly turned to social media to post about their experiences and intents online. 
In some situations, when they experience the {\em loss} of a loved one, their online expression of emotion may evolve into {\em aggression} towards rival gangs
and ultimately into real-world violence. In this paper, we present a novel system for detecting Aggression and Loss in social media. Our system features the use of domain-specific resources automatically derived from a large unlabeled corpus, and contextual representations of the emotional and semantic content of the user's recent tweets as well as their interactions with other users. Incorporating context in our Convolutional Neural Network (CNN) leads to a significant improvement.
\end{abstract}

\section{Introduction}
\kmnote{FROZEN. If you want changes, use a comment.}
\eanote{Adding Desmond's feedback with comments, "D:"}
\fnote{but then it looks so sad!}


In cities such as Chicago, gang-involved youth 
have increasingly turned to social media to post about their experience, often expressing grief when friends or family members are shot and killed. As grief turns to anger, their posts turn to retribution and ultimately to 
plans for revenge~\cite{Pattonetal2018}. Research in this space has shown that online posts often affect life in the real world~\cite{moule2013internet_banging,patton2013chb,pyrooz2015justice_quarterly,patton2016sticks,patton2017tweets}. 
In some communities, violence outreach workers manually scour online spaces to identify such possibilities and intervene to diffuse situations. A tool that identifies Aggression or Loss posts could help them filter irrelevant posts, but resources to develop a tool like this are scarce.

In this paper, we present automatic approaches for constructing resources and context features in this domain, and apply them to detecting Aggression and Loss in the social media posts of gang-involved youth in Chicago. We exploit both a small labeled dataset (4,936 posts) and a much larger unlabeled dataset
(approximately 1 million posts),
\kmnote{To discuss: will we make the data available?}
 which we constructed using a method that enabled us to gather Twitter posts representative of the community we study. We incorporate our approaches into a CNN system, as well as a Support Vector Machine (SVM) to match the architecture of prior work, thus enabling analysis of the impact in different frameworks\footnote{We will make tweet IDs for the data available to researchers who sign an MOU specifying their intended use of the data and their agreement with our ethical guidelines. Contact Serina Chang (sc3003@columbia.edu) or Kathleen McKeown (kathy@cs.columbia.edu).  Our code is available at https://github.com/serinachang5/contextifier.}.

Key features of our system are the use of domain-specific word embeddings and a lexicon automatically induced from our unlabeled dataset. When classifying an individual tweet, our system considers the content and emotional impact of the tweets in the author's recent history.
If applicable, our system additionally takes into account a 
model of
the pairwise interactions between the author and other users in the tweet referenced via either retweet or mention.

We compare our approaches with previous work that used a smaller dataset (800 tweets) and hand-curated resources with an SVM ~\cite{Blevinsetal2016}.
By integrating our induced domain-specific and context information in a CNN, we achieve a significant increase over their reported results. 


Our contributions include:
\begin{itemize}
    \item A new labeled dataset, six times larger than that of prior work;
	\item Domain-specific resources, automatically induced from our constructed unlabeled dataset;
    \item Context features that capture semantic and emotion content in the user's recent posts as well as their interactions with other users in the dataset.
\end{itemize}
  Our approach brings us one step closer to building a useful tool that can help reduce gang violence in urban neighborhoods. 
 In the remainder of the paper, we present related work, the dataset that we used, and our methodology. We conclude with an error analysis and a discussion of the impact of our contributions.

\section{Related Work}
\kmnote{WE MAY WANT TO CUT MORE FROM HERE.}
\eanote{Adding Desmond's feedback with comments, "D:"}
Researchers have begun to explore how online data can be used to help prevent gun violence. \citealt{PavlickAndCallisonBurch-2016:EMNLP:GVDB} are 
creating the Gun Violence Data Base by crowdsourcing annotations on newspaper articles that report on gun violence, labeling the sections of text that report on incidents, the shooter, and the victim.
Researchers have also explored identifying deaths from police shootings with semi-supervised methods for both CNNs and logistic regression~\cite{Keith2017IdentifyingCK} and found that logistic regression using a soft-labeling approach gave the best results. Researchers studying gun control issues analyzed social media for posts related to any issue around guns in the year following the Sandy Hook elementary school shooting \cite{Benton:2016rm} and argued
that online media can be used to understand trends in gun violence and gun-related behaviors~\cite{Ayers:2016uo}.

 Closely related research aims to automatically identify gang members' Twitter profiles ~\cite{Balasuriyaetal2016}. After collecting profiles using bootstrapping, they trained different classifiers on the tweets and meta-information about the authors. Further research analyzes the social networks of gangs~\cite{radil2010spatializing_gangs} and predicts gang affiliation based on the analysis of graffiti style features~\cite{carolyn_rose2014graffiti}.
 
 The most relevant work in automatically analyzing social media posts by gang-involved youth is that of \citealt{Blevinsetal2016}. The labeled dataset that Blevins and collaborators used is extremely challenging, in part due to its size, but also because it contains text in a particular dialect of English -- African American English (AAE) -- which has very little core NLP tool support. Other research investigating the development of tools for understanding AAE in social media~\cite{blodgett-green-oconnor:2016:EMNLP2016} shows that existing tools (e.g., dependency parsers) perform poorly on this language. \svnote{The next statement gives me a feeling that Patton is our collaborator. Can we make the statement more blind} Previous work by Patton on a subset of our dataset notes that due to the linguistic style, tweets from gang-involved youth in Chicago 
can be challenging for outsiders to interpret and thus are often open to misinterpretation and potential criminalization 
\cite{patton2017god}.
 
 The challenges of interpreting our data are further compounded by the usual difficulties with Twitter data. Twitter data is sometimes handled by translating it to Standard American English (SAE) through the use of a phrasebook. The NoSlang Slang Translator \cite{noslang}, and the accompanying NoSlang Drug Slang Translator \cite{noslangdrug}, have been used in other tasks to translate social media communication \cite{sarker2016social}, \cite{han2011twitter}.
 
 To engineer features for an SVM classifier, \citealt{Blevinsetal2016} learned a part-of-speech (POS) tagger for their data and constructed a word level translation phrasebook to map emojis and slang to the Dictionary of Affect in Language (DAL) in order to identify their emotion. 

 In contrast to Blevins' translation approach, we leverage our large unlabeled dataset to automatically induce resources, such as word embeddings, that function well within the domain of our task. Previous research on domain-specific word embeddings includes work in cybersecurity \cite{roy2017cybersecurity}, disease surveillance \cite{ghosh2016disease}, and construction \cite{tixier2016construction}. These domain-specific word embeddings tend to improve performance on tasks within that domain. 

 Context has been used in previous research on detecting hate speech in social media. \citealt{qian2018learning} found significant improvements by collecting the entire history of a user's tweets and feeding them to a encoder to create an intra-user representation, which was used as input to a Bidirectional LSTM. They also used a representation of tweets similar to the tweet being classified. While their approach captures a user profile based on everything the user has posted, in our approach we investigate how the recent history of tweets and interactions with others 
 can improve classification. Others also make use of a user profile, though not one learned from unlabeled data~\citep{dadvaretal2013}. 


\section{Data}

\label{data}
\kmnote{Waiting on Siddharth. Could consider cuts here. Else frozen.}
\eanote{Adding Desmond's as comments suggestions with "D: "}
Our dataset consists of two parts: first, a collection of 
4,936 tweets authored or retweeted by Gakirah Barnes, a powerful female Chicago gang member, 
and her top communicators, as well as additional Twitter users in the same demographic, 
\kmnote{I am removing reference to social work researchers on our team. That may yield our identity. I will leave social work though as it indicates level of expertise. }
annotated by social work researchers 
who have been studying Gakirah and the associated Chicago gangs. Second, we use a much larger collection of approximately one million unlabeled tweets automatically scraped from 279 users in the same social network. This social network is comprised of 214 users snowball-sampled from Gakirah Barnes' top 14 communicators. 
\kmnote{Siddharth will check how many users. This doesn't add up.}
Traditionally, snowball sampling has been used to recruit hard-to-reach research subjects~\cite{Atkinson&Flint2001} and we have adapted it for social media. The remaining 65 users were added to this network by retaining those with the highest IQI score \footnote{\url{https://www.brookings.edu/wp-content/uploads/2016/06/isis_twitter_census_berger_morgan.pdf}} from the full list of Gakirah's Twitter followers. 
Our tweets thus form a representative sample of Twitter dialogue between youth from Chicago neighborhoods with high levels of gang activity during that time period.

The social work researchers performed a detailed, qualitative analysis of a subset of the dataset, with a focus on analyzing how context influences determination of a label. For example, they note that an aggressive tweet may reference a previous event, and will often use coded language to do so. Since much of the language used in our data differs significantly from standard American English, local youth active in similar environments served as consultants to answer questions about the language, as they were able to interpret the slang terms present in these tweets. The social work researchers conducted 
a fine-grained analysis using an online tool for annotation, identifying insults, threats, bragging, hypervigilance and challenges to authority, all of which were collapsed into a single category, Aggression. Posts including distress, sadness, loneliness and death were collapsed into the category Loss. The Other category includes discussion of other aspects of their life, such as friendships, relationships, drugs, general conversations, and happiness. We developed our system (as did \citealt{Blevinsetal2016}) on the collapsed labels, as the task is difficult even with three-way categorization.

Each tweet in a subset of the entire dataset consisting of 3,000 tweets was reviewed by two different annotators. Inter-rater reliability between raters was tracked, with dissimilar annotations flagged for further review. Flagged tweets were further analyzed by the social work researchers, which included youth from Chicago who currently live in the same community as, or an adjacent one to, that in which the deceased Gakirah Barnes resided, 
to adjudicate disagreement.
Among the set of tweets coded by two annotators, inter-annotator agreement on the Aggression class was high even before adjudication, with a Cohen's kappa coefficient of .94; agreement on the Loss class was somewhat lower, with a Cohen's kappa of .83. 
Examples of labeled Twitter posts from Gakirah and her followers are shown in Table~\ref{example-table-2}~\footnote{Our data was scraped from publicly available posts and was determined exempt by our organization's IRB. User names are replaced with USER in the table, and text has been modified to render tweets unsearchable.}.

\fnote{Table has been edited, should be done.}

\begin{table}
    \centering
    \caption{Example tweets}
    \label{example-table-2}
    \begin{tabular}{|p{.3cm}|p{4.7cm}|p{1.2cm}|}
    \hline
    \multicolumn{1}{|l|}{No.} & \multicolumn{1}{c|}{\bf Tweet Text} & \multicolumn{1}{c|}{\bf Label} \\ \hline
    1 & \makecell[l]{\#FreeDaDommmmm [URL] 
    } & Loss \\ \hline
    2 & \makecell[l]{Damn juss peeped shorty on\\
    tha news out here\\
    @USER ..smh..\\
    crazyy.. \#RIPShorty} & Loss \\ \hline
    3 & \makecell[l]{I'm smokin on Dat DMoney\\
   	man Im high as fuck} & Aggress \\ \hline
    4 & \makecell[l]{Lost Ty to Sum Fuck Shit dont \\Fuck around wit Fuck rounds n\\
    u a type of Niggas Ion fuck wit\\
    \includegraphics[height=.8em]{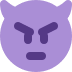}\includegraphics[height=.8em]{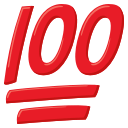}} & Loss \\ \hline
    5 & \makecell[l]{My bro Mooki thirsty he jus\\
    wana \includegraphics[height=.8em]{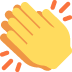} sum \includegraphics[height=.8em]{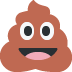} \includegraphics[height=.8em]{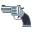} \includegraphics[height=.8em]{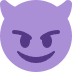} \includegraphics[height=.8em]{100_emoji.png}} & Aggress \\ \hline
 \end{tabular}
\end{table}

In order to mitigate potential issues with training and test data being drawn from different time periods or having different distributions of labels, we shuffled our data and drew stratified samples with equal distribution across classes for our training, validation, and test sets for each of the 
cross validation folds, using 64\%, 16\%, and 20\% of our data for each respectively.
The Aggression and Loss classes are relatively small, reflecting their low distribution in real life: we have only 329 Aggression tweets and 734 Loss tweets, with the Other class comprising the remaining 3,873 tweets.




\section{Methods}
We approach this classification task using a standard CNN classifier 
architecture ~\cite{kim2014convolutional, collobert2011natural} as our starting point. We initially experimented with both character and word level CNNs but found the word level to be 1.6 macro-F1 points better than the character level, so we only include the word level here. 
\kmnote{May also add work on SVM here. Wait until experiments done to see.}
We leveraged the unlabeled corpora by constructing domain-specific embeddings and a lexicon that better fit our unique and low-resource domain. We then integrated our domain-specific resources into the CNN to represent the given tweet as well as to represent context features.
Our context features 
represent a window of the user's recent tweets as well as the interactions of the author with other users via references in their tweets.

\kmnote{Move CNN baseline to Experiments}

\subsection{Domain-Specific Resources}
\label{sect:domainspecific}

We exploited the large unlabeled corpus to build two domain-specific resources for this task: word embeddings and a task-specific lexicon. 
\subsubsection{Word Embeddings}
\label{word_embeddings}
 Word embeddings have proven useful in representing the semantic content of sentences. 
 The semantic representation of a word by its associated embedding, however, depends on its usage in the corpus the embedding was trained on, and so off-the-shelf word embeddings do not always adapt well to tasks with a unique domain \cite{roy2017cybersecurity}, \cite{ghosh2016disease}, \cite{tixier2016construction}.
Thus, we were motivated to use our unlabeled corpus 
to create domain-specific word embeddings. We used the Word2Vec \cite{word2vec} CBOW model to train the embeddings which is the default training algorithm available in Gensim \footnote{\url{https://radimrehurek.com/gensim/models/word2vec.html}}. 
We used a window size of 5 words with a minimum word count of 5 to train $w \in \mathbb{R}^{300}$. The CBOW model was trained for 20 epochs.


\subsubsection{Computing a Lexicon of Aggression and Loss} 
\label{splex}
Given the domain-specific nature of our users' language, we could not rely on standard NLP lexicons to represent emotion in their tweets. For our task, the two emotions of interest are Aggression and Loss. Previous work~\citep{Blevinsetal2016} used a phrasebook to translate the domain-specific words of their corpus to Standard American English so that they could access emotion in the Dictionary of Affect in Language (DAL)~\citep{whissell2009}, but this approach does not generalize to capture new words.

We therefore adapted the SENTPROP algorithm ~\citep{hamilton2016sentprop} to automatically induce a lexicon of Aggression and Loss from our unlabeled corpus. The SENTPROP algorithm constructs a lexical graph out of the word embeddings, then propagates labels from the seed sets over the unlabeled nodes via a random walk method. The resulting output for each word indicates the probability of a random walk from the seed set landing on that node. We chose SENTPROP as an induction method because it performs especially well for domain-specific corpora, and it is resource-light and interpretable.

We created word embeddings by employing an SVD-based method that was reported by the SENTPROP authors to perform optimally with their algorithm. We first constructed the positive point-wise mutual information matrix, $M^{PPMI}$, over the unlabeled corpus, then computed singular value decomposition (SVD) to derive $M^{PPMI}$ = $U\Sigma V^{\intercal}$. 
The word embedding for word $w_i$ was thus given by $U_i$, truncated to a standard length of dimension 300. To construct our seed sets, we asked our annotators to consider words for Loss and Aggression which they associated most strongly with each class. They generated a set of 29 words for Aggression and a set of 40 words for Loss, which we include in our appendix.

We ran SENTPROP with our SVD-based embeddings and the seed sets from our annotators. We used the output probabilities from the random walks to map words to their association with Aggression and Loss, thus forming our lexicon of Aggression and Loss. Finally, we scaled the probabilities per class to mean 0 and variance 1.

\subsection{Context Features}


Our context features utilize the domain-specific resources that we induced from the unlabeled corpora. To capture context, we first considered the author's recent history, separately exploring representations by our domain-specific word embeddings and by the SENTPROP lexicon (SPLex). If applicable, we also considered the interactions between the author and other users who were referenced in the tweet, either via retweet or mention.

\subsubsection{User History}
To obtain the user's recent history, we ordered all the tweets chronologically and bucketed them by author. Thus, for any given tweet occurring at time t, $a_t$, we were able to retrieve previous tweets $a_{t-1}, a_{t-2}, \ldots$ by that user. We treated recent history as a sliding window and fetched tweets within the past $d$ days from when the current tweet was tweeted, such that recent history tweets would be the set \{$a_{t-1}, \ldots, a_{t-k}$\}, where $t-k < d$. 

To represent the tweets within the context of recent history, we first combined word level representations into tweet level, then tweet level representations into context level. At each stage of combination, we tried both summing and averaging. Thus, our recent history representations were built by aggregating either word embeddings or SPLex scores, which maintained their dimensionality of
300 or 2, respectively.

We also considered three types of tweets that would be relevant to a user. The user's own tweets (\texttt{SELF}) would always be relevant; we experimented with also including tweets where the user was retweeted (\texttt{RETWEET}) and tweets where the user was mentioned (\texttt{MENTION}). We included these parameters as additional sources of context 
because a user's tweet may be a response to a recent mention or retweet from another user.

We also experimented with weighting the most recent tweets more heavily than further tweets within the recent history window. This became especially important when we experimented with larger windows of a month or more, since tweets from a few days ago are more likely to be related to the current tweet than tweets from a few weeks ago. To model this diminishing relevance, we introduced a weighting protocol with a variable half-life where weights decay exponentially over time. The parameter we tuned was the half-life ratio $r$, 
which is the proportion of the window size $d$ 
that corresponds to the window's half-life. Then, before combining tweet level representations into context level, we multiplied each tweet representation $b_i$ by its weight, $2^{-\frac{\Delta t}{f}}$, where $\Delta t = t-i$ is the distance in days between the context tweet $a_i$ and the current tweet $a_t$, and $f = d*r$ is the half life.

\subsubsection{User Interactions}
As an additional context feature, we modeled the pairwise interactions between users. To identify interactions, we iterated through our unlabeled and labeled corpora and checked which users were involved in each tweet. We counted a user as involved in a tweet if they posted the tweet or were referenced via retweet or mention. For each pair of users, we aggregated all their tweets of mutual involvement into one document and averaged the document's word embeddings to create a representation of their pairwise interactions in $\mathbb{R}^{300}$.

\section{Experiments}

We experimented with the efficacy of our domain-specific resources, the impact of different context parameters, and the contribution of context to predicting Aggression and Loss.

\subsection{Corpus pre-processing}
 For word level models, we preprocess each tweet by: i)~lowercasing every character, 
 ii)~replacing every user mention and url with special tokens ``user" and ``url",
iii)~considering each emoji an individual token, whether space separated or not,  and iv)~removing emoji modifiers to reduce sparsity, just as we used lowercasing. We 
select the top 40K tokens based on frequency, replacing the remaining tokens with ``UNKNOWN". We zero-pad or trim tweets so that tweet length will be 50 when passed to our CNN model. Similarly, we only consider users who occur (as author, source of retweet, or in mention) in the labeled and unlabeled corpus at least twice, resulting in 35,656 users in total.

We extract the author of the tweets from meta-data, and user mentions and original posters of retweets from the Twitter text, based on their Twitter display name. We used Twitter display name rather than user ID because we cannot collect user ID for interaction features.

\subsection{CNN Architecture}
For this 3-way classification task, we train two models; the first model predicts whether a tweet has the Aggression label and the second predicts for Loss. Each model maps a sequence of tokens to a probability value for a class. Here we define the architecture of our CNN model. Our input $c$ is a token index sequence of length 50. We map each token index to a vector $\in \mathbb{R}^{300}$ with a trainable embedding matrix, followed by dropout 0.5. We apply a 1D Convolutional layer with kernel sizes 1 and 2, filter size 200 each, to the embedded token sequence, followed by ReLU activation, max pooling and dropout 0.5. We concatenate the output of max pooling for kernel sizes 1 and 2, stack another dense layer $h$ with dimension 256, and connect the output of $h$ to the final single output unit with sigmoid activation.

In the prediction phase, for each data point, we classify it as Aggression if the the first model produces the probability score above threshold $t_{A}$. If it is not predicted as Aggression, then we classify it as Loss if the second model produces a score above a threshold $t_{L}$. The remaining tweets are classified as Other. $t_{A}$ and $t_{L}$ are tuned on the validation set.

We incorporate context information into the neural network in the following way. Each type of context feature takes the form of a real vector: both word embedding user history and word embedding user interaction features are in $\mathbb{R}^{300}$, and SPLex user history features are in $\mathbb{R}^{2}$. We concatenate these feature vectors with the last layer $h$ before the final classification output.

\subsection{SVM Baseline}
\kmnote{Fei-Tzin to update this section to indicate what changed} \fnote{done, i think}
We used as our baseline method a linear-kernel SVM classifier as used by \citealt{Blevinsetal2016}.
We obtained code from the authors and trained on our larger dataset.
In this method, after basic preprocessing is performed to replace urls and user mentions with special tokens, unigram, bigram, part-of-speech tag, and emotion features are extracted.
Feature selection is performed to prune the feature space.
The part-of-speech tagger used in \citealt{Blevinsetal2016} was developed for use on this domain; emotion features are computed using scores for each tweet word taken from the Dictionary of Affect in Language (DAL). 
We performed gridsearch to re-tune the loss function, the regularization penalty type, and the penalty parameter C,
but found that the original settings for these parameters still performed best even on our new development set. We also tuned the class weights used: while the model performed best on the original data with balanced class weights, we found that less extreme balancing performed better here (weights 2, 1, and 0.12 for Aggression, Loss, and Other, respectively).

While we retrained the SVM on our new training set, we did not modify the additional components used for feature selection such as the phrase table or the specialized part-of-speech tagger, as we had no additional data available for this. This indicates the difficulty of generalizing to new data with unseen vocabulary, and is one of the disadvantages of using manually-created specialized feature sets such as these.

\subsection{Domain Experiments}
In order to test the efficacy of our domain-specific word embeddings, we compared them with a number of other embedding types. Our baseline method was \citealt{glove}'s GloVe embeddings pretrained on a general Twitter dataset, available from their website\footnote{\url{https://nlp.stanford.edu/projects/glove/}}. We trained a parallel set of word embeddings on the African American English (AAE) corpus of around 1.1 million tweets provided by \citealt{blodgett-green-oconnor:2016:EMNLP2016}, and another set on a corpus of a location-specific set of tweets that we scraped, drawn from users who posted from a specific area within the South Side of Chicago where the gangs we study are based. 
We also compared performance with a randomly initialized word embedding matrix.

\subsection{Context Experiments}



We first explored the impact of the user history parameters, tuning them separately for representations by our domain-specific word embeddings 
and by SPLex. We kept these representations separate because we expected them to capture different types of context: word embeddings should capture the semantic content of the user's history, while SPLex scores should capture something closer to the user's emotional state leading up to the tweet.

With each representation, we experimented with summing versus averaging word embeddings to yield a tweet level representation, and similarly experimented with summing and averaging from tweet embeddings to context level representations. We varied the size of the context window, $d$, trying 2 days, 1 week, 1 month, 2 months, and 3 months. We also varied the half-life ratio, $r$ = .25, .5, .75, or no weighting. Lastly, we tried including different types of posts in the user history.

Once we tuned the user history parameters, we experimented with adding our context features (user history and user interactions) to the best tweet level model we could achieve without context. For our CNN, our best tweet level model used our domain-specific word embeddings as pretrained weights for the embedding layer (CNN-DS in Table 3). To evaluate the impact of our resources in different frameworks, we additionally experimented with the contribution of context in an SVM. The best tweet level SVM included the averaged domain-specific word embeddings and summed SPLex scores of the tokens in the tweet (SVM-DS).

\section{Results and Discussion}
\begin{figure*}[htbp]
\centering
\caption{Diagram of our steps to generate domain-specific and context features for our neural net system.}
\label{fig:architecture}
\includegraphics[width=\textwidth]{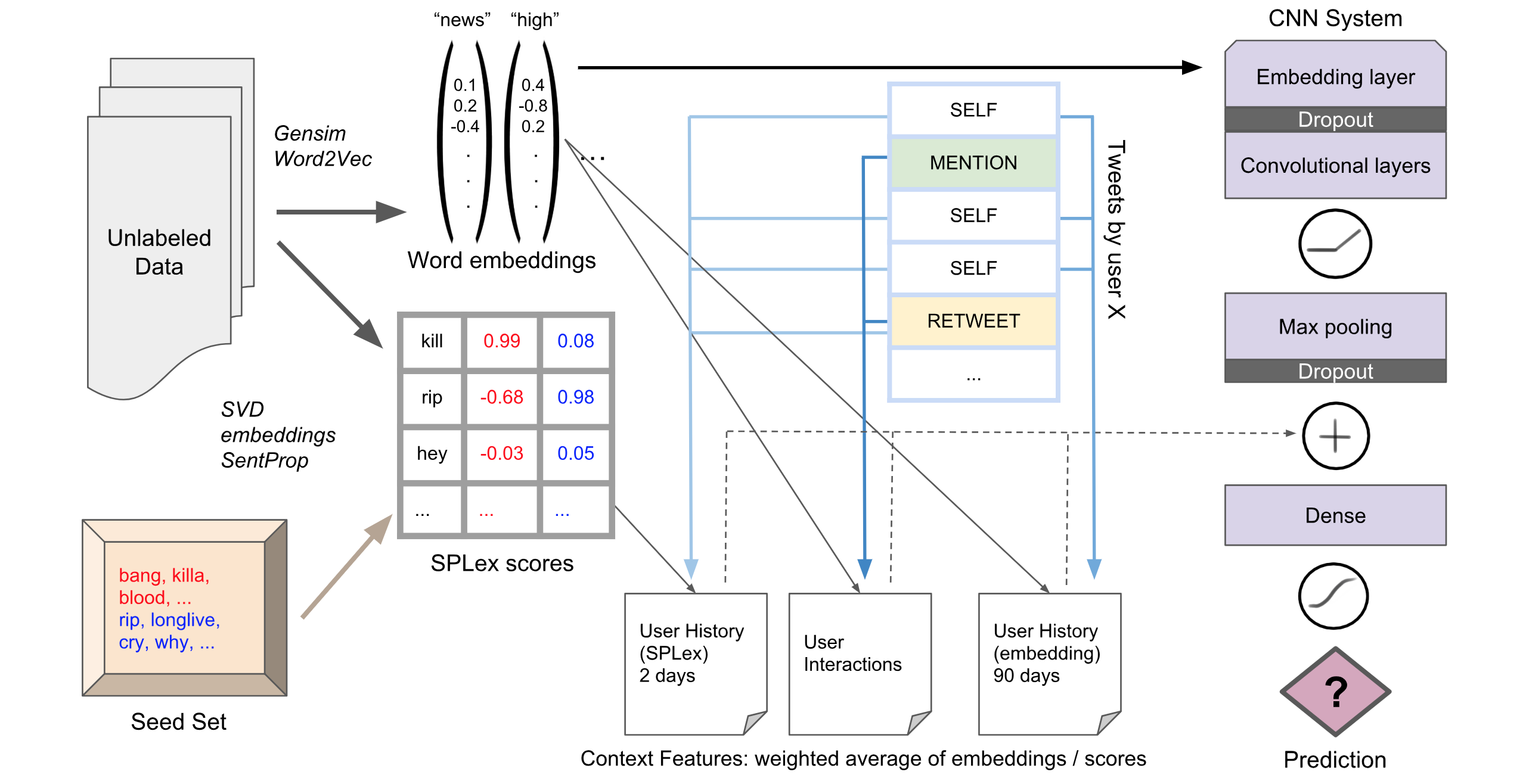}
\end{figure*}
We report results comparing different embeddings and comparing parameters for context. We use the best results from these experiments to produce our final systems in the SVM and CNN frameworks. The best resulting architecture for the CNN framework is illustrated in Fig.~\ref{fig:architecture}.


\subsection{Comparison of Embeddings}
Experiments were performed using five-fold cross-validation over the labeled data and were repeated five times for each fold to account for variance between runs.
Reported F-scores, shown in Table \ref{table:embedding}, are averaged across runs and across folds. 

Word embeddings trained on our unlabeled corpus outperformed other embeddings by over 4 points. Related datasets such as the location-specific or AAE corpus did not provide helpful semantic information, as their embeddings did not even beat random initialization. This was not an effect of corpus size, since these corpora contained 800,000 and 1.1 million tweets, respectively, compared to the 1 million in our unlabeled corpus. Thus, we attribute the difference to the importance of deriving embeddings directly from our community of interest, demonstrating that the language of our community is more specific than AAE in general and that our snowballing method was able to capture a better representation of user language than a location driven method. 

\begin{table}
\centering
\caption{Results comparing different embeddings with CNN. GN refers to Google News, LS to location specific embeddings, GT to Glove Twitter embeddings, and DS to our domain specific embeddings. A, L and O refer to Aggression, Loss, and Other respectively.}
\label{table:embedding}
\begin{tabular}{|l|c|c|c|c|}
\hline
\multicolumn{1}{|c|}{\begin{tabular}[c]{@{}c@{}}Embeddings\\ Type\end{tabular}} & \multicolumn{3}{c|}{\begin{tabular}[c]{@{}c@{}}F1 \end{tabular}} & {\begin{tabular}[c]{@{}c@{}}Macro\\F1\end{tabular}}      \\ \hline
\multicolumn{1}{|c|}{}                   & A              & L                    & O                   &               \\ \hline
GN                              & 27.9          & 66.6           & 86.9           & 60.5 \\ \hline
AAE                                     & 27.3           & 69.8           & 86.5           & 61.2 \\ \hline
LS                        & 31.3           & 68.3           & 87.9          & 62.5 \\ \hline
Random Init.                             &  29.3        & 70.5          & 88.9           & 62.9 \\ \hline
GT                            &  29.0          & 71.1         & 89.0           & 63.0 \\ \hline
DS                          & \textbf{37.9}           & \textbf{73.4}           & \textbf{90.3}           & \textbf{67.20} \\ \hline
\end{tabular}
\end{table}

\subsection{User History Parameters}
Experiments were performed using five-fold cross-validation and F-scores computed as in the word embedding experiments. We found that user history represented by domain-specific word embeddings performed optimally when we averaged from word to tweet level and from tweet to context level. The best window size was $d = 90$ days, including only \texttt{SELF} posts, and using a half-life ratio of $r = 0.25$. For user history represented by SPLex, we found the best method of combination to be summing, at both the word and tweet level. We hypothesize this is because summing captures not only the presence but also the number or density of highly indicative Aggression or Loss words posted by the user over the context window. The best window size was $d = 2$ days, including both \texttt{SELF} and \texttt{RETWEET} posts, without half-life weighting.

\begin{table*}[htbp]
\centering
\caption{Comparison of different models. The below pairs of algorithms achieve statistical significance $p < 0.002$ for each class (the higher performing algorithm comes first): i) CNN-Context vs. CNN-DS; ii) CNN-DS vs. SVM-Retrained; iii) SVM-Context vs. SVM-DS. SVM-Context outperforms SVM-Retrained in the Aggression class by a robust margin (5 points).}
\label{table:bestsystem}
\begin{tabular}{|l|c|c|c|c|c|c|c|c|c|c|}
\hline
Model                & \multicolumn{3}{c|}{Aggression} & \multicolumn{3}{c|}{Loss} & \multicolumn{3}{c|}{Other} & Macro F1              \\ \hline
                     & P         & R        & F        & P       & R      & F      & P       & R       & F      & \multicolumn{1}{l|}{} \\ \hline
SVM-Retrained(baseline)        & 36.4      & 31.3     & 33.7     & 73.7    & 68.8   & 71.2  & 89.8    & \textbf{92.0}    & 90.9   & 65.3                  \\ \hline
SVM-DS               &  32.4     & 38.9     & 35.4     & 66.9    & 72.9   & 69.8   & 90.8    & 87.7    & 89.2   & 64.8                  \\ \hline
SVM-Context          & 35.0      & 43.7     & 38.8     & 68.6    & 74.0   & 71.2   & \textbf{91.6}    & 88.2   & 89.9   & 66.6                  \\ \hline
CNN-DS               & 35.7      & 41.1     & 38.2     & \textbf{78.9}    & 70.3   & 74.3   & 90.7    & 91.4    & 91.0   & 67.9                  \\ \hline
CNN-Context          & \textbf{38.3}  & \textbf{46.4}   & \textbf{42.0}      & 78.8     & \textbf{73.2}     & \textbf{75.9}    & 91.3   & 91.7   & \textbf{91.5}    & \textbf{69.8}                   \\ \hline
\end{tabular}
\end{table*}

Our approach was designed to implement and test previous insights about the domain, particularly that context plays a role in the interpretation of posts. The short time frame for SPLex user history corresponds with the 2 day window found in \citealt{Pattonetal2018}'s 
research and reflects the fact that emotional states may fluctuate often and within a certain number of days. 
In contrast, word embeddings improved consistently as we extended the context window from 2 days to 90 days. Since word embedding user history is meant to capture the user's semantics, a larger window size means the representation can be drawn from more tweets, and thus reflects a more representative sample of the user's semantics around this time period.

\subsection{Comparison of Best Systems}
To develop a more stable measurement of comparison between different systems, we create four independent sets of 5-fold cross validation splits on our data set (altogether 20 folds); to account for randomness in neural net training, we train each neural net model 5 times and take the majority vote of the predictions. For each class, we calculate the statistical significance of F-score based on the predictions on the concatenated test sets of all 20 folds using the Approximate Randomization Test (\citealt{riezler2005some}) with the Bonferroni correction
for multiple comparisons. Results are shown in Table \ref{table:bestsystem}.

Adding context contributed to a significant improvement in both the CNN and SVM frameworks, demonstrating the independent value of our context features over domain-specific resources. For contrast, we also compared our context features with user profiles built from averaging the word embeddings in all of the user's tweets. Our pairwise and user history features outperformed user profiles by .7 points, demonstrating that it is valuable to provide dynamic representations of users that can adjust to their recent posts or their interactions with other users, as opposed to stereotyping their overall behavior.

Additionally, we compare the impact of our domain-specific resources to those used by \citet{Blevinsetal2016}. In particular, we expect that their emotion scores will not generalize to the new vocabulary in our large unlabeled corpus (see Section~\ref{sect:domainspecific}).
Our domain-specific resources alone without context raise our SVM to comparable performance with the Blevins et al. retrained baseline, and the resources push our CNN without context over this baseline. This demonstrates that our automatic methods can do as well as if not better than phrasebook methods, and they are significantly more efficient to generate.

\section{Error Analysis}

In this section we provide an analysis of the trade-offs of each classifier by analyzing some of the examples in Table \ref{example-table-2}.


{\bf Context vs non-context CNN.} Our best CNN - a system which incorporated context - was able to correctly predict tweets 3 and 4, whereas our baseline using only our pretrained Word2Vec embeddings was not. Correctly classifying tweet 4 relies on the knowledge that the referenced user, DMoney, is a deceased member of a rival gang of the poster. In tweet 3, the poster is saying that he has seen Gakirah's death on the news; this is an expression of loss. 

{\bf Domain-specific vocabulary.} Our CNN trained on domain-specific word embeddings is able to correctly classify tweet 5, while the one trained on Twitter word embeddings did not pick up the aggressive content. This user is talking about how their friend is ready to kill someone.  This tweet contains the word {\em thirsty} but in this domain-specific context it means being ready and having an urge (although it would not always refer to killing).

{\bf Hashtags and character sequences.} 
Despite their strengths, both our best CNN and our best SVM classifiers were still unable to correctly classify some of the trickier cases. There were certain types of tweets they were categorically unable to recognize: tweet 1 features a hashtag that refers to an incarcerated acquaintance of the poster, but as both our CNN and SVM models operate at the word level, this tag would have appeared simply as a rare or unknown token to them.

{\bf Anger miscategorized as Aggression.} At times, the classifier categorized posts that express anger as Aggression. For example, in tweet 4 the author uses profanity to express grief related to the loss of a friend. In addition, the devil face emoji, which is sometimes used to express aggression, is also used in the context of anger. 
While the best CNN model managed to correctly predict this as Loss, the SVM miscategorized it as Aggression.

\section{Ethics}

Our ethics guidelines include just treatment of the users who provide our data, removal of identifying information for 
publication, and the inclusion of Chicago-based community members as domain experts in the analysis and validation of our findings. 

There are risks involved with detecting Aggression and Loss in social media data using automatic detection systems. These risks include possible misidentifications of tweets, increased police involvement, and loss of privacy, which all have the potential to harm marginalized communities and people. Our mitigation strategies begin by partnering with violence prevention organizations and incorporating domain experts~\citep{Freyetal2018} to ensure the highest ethical standards for interpreting social media posts and for the dissemination and use of our research for violence prevention. Through insights gained from these partnerships, we developed our own risk mitigation strategies: de-identifying each tweet and rendering it unsearchable through textual modification without altering meaning; encrypting our social media corpus to protect user identities; and relying on violence prevention organizations’ expertise in deciding if and when to involve law enforcement to prevent the unethical use of our data (e.g., hyper-surveillance of communities of color).

\section{Conclusion and Future Work}


Our approach shows that integrating emotions and semantic content of a user's recent posts is an important component for the task of predicting Aggression and Loss in social media posts of gang-involved youth. 
Furthermore, using domain-specific embeddings and an Aggression-Loss lexicon induced from a corpus of language constructed to represent our specific community of users is also critical to success. Our experiments reveal that our snowballing technique  is more effective than a location based approach and that fitting our community is more complex than resorting to their demographic, as captured in the AAE corpus of \citet{blodgett-green-oconnor:2016:EMNLP2016}.

Our work has real life implications for the use of machine learning to identify unique characteristics in social media data that may indicate the process by which gun violence may occur~\citep{Pattonetal2018B}. Our partnership between computer scientists, social work researchers and practitioners has advanced plans to create applications to help outreach workers in Chicago identify factors related to potential violence, potentially allowing them to prevent and intervene in aggressive online activity. The tool, which would be co-created with community stakeholders, would enable quick scanning of large quantities of social media posts that outreach workers would be unable to perform manually. 

We expect our methods to be generalizable because we compute embeddings and lexicons from neighborhood-specific data and do not rely on large, hand-crafted resources such as dictionaries. However, we hope to test generalizability in future work by applying our methods to other gang-related corpora, because there is variation in language, local concepts, and behavior across gangs. In the future, we are also interested in further experimenting with the context features introduced in this work; for instance, by extending our pairwise interaction features to take into account direction between users. Finally, we intend to explore other types of context, such as reference to specific events that may trigger the emotions of either Aggression or Loss.

\section{Acknowledgements}
This research is supported in part by DARPA contract 55630053. The authors also thank the anonymous reviewers for their thoughtful comments.

\bibliography{emnlp2018}
\bibliographystyle{acl_natbib}

\newpage

~

\newpage
\section{Supplemental Material}
\label{sec:supplemental}

In this section, we include specific details of our Aggression and Loss lexicon, which we refer to as SPLex.

\subsection{Seed Sets}
This table lists the full seed sets generated by our annotators for Aggression and Loss. 
\begin{table}[H]
\centering
\label{my-label}
\begin{tabularx}{\linewidth}{|c|X|}
\hline
Aggress & `angry', `opps', `opp', `fu', `fuck', `bitch', `smoke', `pipe', `glock', `play', `missin', `bang', `smack', `slap', `beat', `blood', `bust', `bussin', `heat', `BDK', `GDK', `snitch', `cappin', `killa', `kill', `hitta', `hittas', `shooter', `tf' \\ \hline
Loss  & `free', `rip', `longlive', `LL', `rest', `up', `restup', `crying', `cry', `fly', `flyhigh', `fallin', `bip', `day', `why', `funeral', `sleep', `miss', `king', `hurt', `gone', `cant', `believe', `death', `dead', `died', `lost', `killed', `grave', `damn', `soldier', `soldiers', `gang', `bro', `man', `hitta', `jail', `blood', `heaven', `home' \\ \hline
\end{tabularx}
\caption{Seed sets for Aggression and Loss}
\end{table}


\end{document}



\section{Supplemental Material}
\label{sec:supplemental}

In this section, we first cover details of Convolutional Neural Networks (CNNs) and Multi-Layer Perceptrons (MLPs) for completeness. Next we list the implementation specific details of our models which could not be covered in the main sections. Our anonymized code is open source and is available at \url{https://goo.gl/CvHdaT}.

\input{Methods_Defs.tex}

\subsection{Text Preprocessing}
We perform some text preprocessing on the raw tweets like replacing urls,user mentions and retweet markers with special tokens. We keep the hashtags intact. For word level CNNs and SVM, we use nltk tokenizer to tokenize the tweets. We use utf-8 encoding to handle the emojis and non-ascii characters. Thus our character set turned out to be around 1700 unique characters.

\subsection{Baseline CNN}
\label{sec:supplemental_baseline_cnn}


Below we list the various hyperparameters.

\begin{itemize}
    \item Our baseline CNN model was implemented in Keras deep learning framework
    \item Choice of Optimizer: We used the Nadam optimizer available in Keras with default learning rate
    \item Filter widths: $1, 2, \ldots, 5,$
    \item Number of filters per width: $F = 200$ 
    \item Hidden layer dimension: $H = 256$
    \item Dropout: 50\% after the embedding layer and convolutional layers
    \item Weight initialization: default weight initialization available in Keras except when using pre-trained weights
    \item Early stopping : Validation macro-F score was used as a criterion for early stopping. Keras does not support this. we implemented our own callback to track macro-F score.
\end{itemize}

We varied the filter widths by dropping filters of size 4 \& 5, however it let to some degradation. Similarly dropping filter of size 1 also lead to degradation indicating that unigrams especially emojis are important. We also varied the number of feature maps and obtained best results on validation set with 200. We did not tune other hyperparameters.

\subsection{CNN With Distant Supervision}
The hyperparameters used to trained CNN on distantly labeled data were same as above.

\subsection{CFP}

\begin{itemize}
    \item Our CFP model was implemented in Tensorflow. We could not use Keras because it does not offer enough flexibility.
    \item Choice of Optimizer: The Nadam optimizer available in tensorflow has a bug and does not work as it is. We managed to find the fix on tensorflow's github repository and successfully used it with default parameters.
    \item Filter widths: $1, 2, \ldots, 5,$
    \item Number of filters per width: $F = 200$
    \item Hidden layer dimension: $H = 256$
    \item Dropout: 50\% after the embedding layer and convolutional layers
    \item Weight initialization: We used Xavier initialization for convolutional and dense layer weights.
    \item This model was trained for 12 epochs.
\end{itemize}

In this case, we did not tune the number of positive samples and number of negative samples. Also we did not perform hyperparameter tunning of CNN filter widths, number of filters per width, dropout etc. This is because we wanted to re-use the CNN weights for the supervised task.